\documentclass[letterpaper]{article} 
\usepackage{aaai2026}  
\usepackage{times}  
\usepackage{helvet}  
\usepackage{courier}  
\usepackage[hyphens]{url}  
\usepackage{graphicx} 
\usepackage{natbib}  
\usepackage{caption} 
\usepackage{algorithm}
\usepackage{algorithmic}
\usepackage[table]{xcolor}
\usepackage{xcolor}
\usepackage{bm}
\usepackage{amsmath,amsfonts,amsthm}
\usepackage[font=footnotesize,skip=0pt]{subfig}
\usepackage{multirow}
\usepackage{newfloat}
\usepackage{listings}
\DeclareCaptionStyle{ruled}{labelfont=normalfont,labelsep=colon,strut=off} 
\floatstyle{ruled}
\newfloat{listing}{tb}{lst}{}
\floatname{listing}{Listing}
\title{GRAVITY: A Controversial Graph Representation Learning for Vertex Classification}
\author{
    Etienne Gael Tajeuna\textsuperscript{\rm 1}, 
    Jean Marie Tshimula\textsuperscript{\rm 2,3}
}
\affiliations{
    \textsuperscript{\rm 1}Department of Computer Science and Engineering, Université du Québec en Outaouais, QC J8X 3X7, Canada\\
    \textsuperscript{\rm 2}Mathematics, Statistics and Computer Science Department, University of Kinshasa, Kinshasa, DRC\\
    \textsuperscript{\rm 3}Department of Computer Science, Université de Sherbrooke, QC J1K 2R1, Canada\\
    Email: etiennegael.tajeuna@uqo.ca

}

\usepackage{bibentry}

\begin{document}

\newcommand{\code}[3]{\mathcal{E}\mathit{ncod}(#1\,|\,\bm{\mathcal{#2}},\,\Theta_{\mathcal{#3}})}
\newcommand{\decode}[2]{\mathcal{D}\mathit{iscr}(\mathbf{#1},\, \Theta_{\mathcal{#2}})}
\newcommand{\losscode}[1]{\mathcal{L}_{\mathit{encod}}\Bigl( \Theta_{\mathcal{#1}}\Bigr)}
\newcommand{\lossdiscr}[1]{\mathcal{L}_{\mathit{discr}}\Bigl( \Theta_{\mathcal{#1}}\Bigr)}
\newcommand{\In}[1]{\mathcal{I}\mathit{n}\left(#1\right)}
\newcommand{\Out}[1]{\mathcal{O}\mathit{ut}\left(#1\right)}
\newcommand{\silh}[1]{\mathit{Sil}\left(#1\right)}
\newcommand{\Etiket}{\{1,\,...,\,K\}}

\maketitle

\begin{abstract}
In the quest of accurate vertex classification, we introduce GRAVITY (Graph-based Representation leArning via Vertices’ Interaction TopologY), a framework inspired by physical systems where objects self-organize under attractive forces. GRAVITY models each vertex as exerting influence through learned interactions shaped by structural proximity and attribute similarity. These interactions induce a latent potential field in which vertices move toward energy efficient positions, coalescing around class-consistent attractors and distancing themselves from unrelated groups. Unlike traditional message-passing schemes with static neighborhoods, GRAVITY adaptively modulates the receptive field of each vertex based on a learned force function, enabling dynamic aggregation driven by context. This field-driven organization sharpens class boundaries and promotes semantic coherence within latent clusters. Experiments on real-world benchmarks show that GRAVITY yields competitive embeddings, excelling in both transductive and inductive vertex classification tasks.
\end{abstract}

\section{Introduction}
\subsection{Background}
Graphs are expressive data structures that model pairwise relationships among entities and are extensively used in applications such as social networks, citation networks, and biological systems \cite{newman2003structure, battaglia1806relational}. In many practical settings, graphs are \textit{attributed}, meaning that each vertex is associated with descriptive features, and \textit{homogeneous}, where all nodes and edges are of the same type \cite{kipf2016semi}. Learning effective representations of vertices, commonly referred to as \textit{vertex embeddings} has become a core task in graph-based learning due to its impact on downstream applications such as vertex classification, link prediction, and clustering \cite{hamilton2017representation, wu2020comprehensive}.

Vertex embedding methods can be broadly categorized into \textit{transductive} and \textit{inductive} frameworks. Transductive approaches such as DeepWalk \cite{perozzi2014deepwalk} and node2vec \cite{grover2016node2vec} rely on access to the entire graph during training and are not directly applicable to new vertices. In contrast, inductive methods such as GraphSAGE \cite{hamilton2017inductive} and GAT \cite{velickovic2017graph} leverage local neighborhoods and vertex features to generalize embeddings to unseen vertices, making them more scalable for dynamic or evolving graphs.

\subsection{Motivation}
While many graph neural networks (GNNs) are designed to preserve local proximity and aggregate information from immediate neighborhoods, they often struggle to uncover the latent substructures embedded in the graph topology. To address this issue, various approaches have proposed to enhance representation learning by preserving structural identity or learning modular substructures \cite{you2020graph, dwivedi2023benchmarking}. Some clustering-based methods \cite{tian2014learning, bo2020structural} adopt a two-phase framework: first learning embeddings, then applying clustering algorithms, mirroring the pipeline of spectral methods. However, these models depend heavily on how well the latent space preserves the true organizational patterns of the graph.

Furthermore, many spatial convolution models rely on a fixed $h$-hop aggregation scheme, potentially missing broader topological cues needed to separate structurally similar but distant regions \cite{wu2020comprehensive}. A more dynamic mechanism for interaction and separation of vertex representations is needed to adaptively refine structural context throughout training.

\subsection{Contribution}
To extract meaningful features from attributed homogeneous graphs, we introduce GRAVITY (Graph-based Representation leArning via Vertices' Interaction TopologY), a supervised graph representation learning framework specifically designed for vertex classification. Inspired by physical systems, GRAVITY simulates a dynamic interaction process among vertices, where attractive forces organize the latent space into class-consistent regions.

At the heart of GRAVITY lies a competitive interaction mechanism—a \textit{game of strength}—in which labeled vertices attract others from the same class and repel those from different classes. This dynamic encourages the emergence of well-separated class clusters in the embedding space. We formalize this mechanism through a \textit{gravitational neural network} that learns vertex embeddings by capturing both evolving neighborhood structures and semantic class affinities. 

GRAVITY also introduces an adaptive neighborhood aggregation strategy, enabling each vertex to dynamically adjust its receptive field during training, rather than relying on a fixed $h$-hop scheme. This flexibility enhances the model's ability to preserve both local and global graph structures.

The main contributions of this work is as follows:

\noindent $\bullet$ We propose GRAVITY, a supervised vertex embedding framework based on attractive dynamics, which leverages label information to guide representation learning for vertex classification. 

\noindent $\bullet$ We present a spatial graph convolution mechanism with adaptive neighborhood depth, allowing each vertex to determine its own structural context during learning.

\noindent $\bullet$ We empirically validate the effectiveness of GRAVITY across multiple datasets, demonstrating that it achieves competitive performance in vertex classification by uncovering semantically meaningful and geometrically coherent latent structures.

\section{Preliminaries}
\subsection{Problem statement}
Given $\bm{\mathcal{O}} = \{o_i\}_{i=1}^N$, a set of  $N{>}1$ objects, we use $\bm{\mathcal{X}} = \{x_i\}_{i=1}^N$, $x_i \in \mathbb{R}^d$ ($d \in \mathbb{N}^*$) to denote the corresponding set of attribute vectors representing objects in $\bm{\mathcal{O}}$. According to the context, there exists an ensemble of relationships $\bm{\mathcal{V}} = \{(o_{i},\,o_{j})\,|\,o_{i},\,o_{j}\in \bm{\mathcal{O}}\}$ with its corresponding weights $\bm{\mathcal{W}} = \{\omega_{i, j} \in \mathbb{R} \,|\,\forall (i,\,j),\; (o_{i},\,o_{j})\in \bm{\mathcal{V}}\}$ relating the co-existence of objects $\bm{\mathcal{O}}$. We use $\bm{\mathcal{G}} = \{\bm{\mathcal{O}},\, \bm{\mathcal{X}},\, \bm{\mathcal{V}},\, \bm{\mathcal{W}}\}$ to denote the geometric structure mimicking these overall relationships. Based on the geometric properties of the graph, we assume that it can be partitioned into $K\in \mathbb{N}^*$ subgraphs $\{g_{\kappa}\}_{\kappa = 1}^K$, $g_{\kappa} = \{\bm{\mathcal{O}}_{\kappa},\, \bm{\mathcal{X}}_{\kappa},\, \bm{\mathcal{V}}_{\kappa},\, \bm{\mathcal{W}}_{\kappa}\}$.

The controversial game of strength proposed in this paper consists of mapping attributes vector into a latent space and thus learn novel representation $\bm{\mathcal{Y}} = \{y_i\}_{i=1}^N$ of $\bm{\mathcal{O}}$ where $y_i \in \mathbb{R}^q$, $q<d$. This representation (or embedding) can be written as  $\bm{\mathcal{Y}} = \bigcup_{\kappa = 1}^K \bm{\mathcal{Y}}_{\kappa}$, with $\bm{\mathcal{Y}}_{\kappa}$ the novel representation of vertices of the substructure $g_{\kappa}$.

In the rest of the paper, when going to mathematical operation over the different ensembles, we will use the matrices $\mathbf{X} \in \mathbb{R}^{N\times d}$, $\mathbf{\Omega} \in \mathbb{R}^{N\times N}$, $\mathbf{Y} \in \mathbb{R}^{N\times q}$ to refer to the set of attributes, weights and embedding respectively.

\subsection{Gravitational dynamics}

Gravitational force in classical physics describes the mutual attraction between two masses. According to Newton's law of universal gravitation, the force $f_{i,j}$ between two particles $o_i$ and $o_j$ is derived from a potential function $\Phi$ defined as:
\begin{align}\label{Eq:gravitation-physics}
    f_{i,j} &= \nabla \Phi(o_i, o_j | m_i,m_j, x_i, x_j) \\
    &= \mathbf{W} \cdot \frac{m_i m_j}{\|x_i - x_j\|^2} \cdot \frac{(x_j - x_i)}{\|x_j - x_i\|} \, , \notag
\end{align}
where $\mathbf{W}$ is the gravitational constant, $m_i$ and $m_j$ are the masses, and $x_i, x_j$ denote their positions in space.

In the context of this work, we draw inspiration from this principle to guide vertex interactions during representation learning. Specifically, we treat vertices as particles in a latent space and model an \textit{attractive force} between them when they share the same class label. Meanwhile, vertices of different classes are kept distinct through an opposing influence that discourages closeness.

This analogy provides a conceptual and computational framework to iteratively adjust vertex positions in the embedding space, forming orbit-like, class-aligned clusters. It enables latent dynamics where geometric coherence emerges from local forces, rather than static neighborhood aggregation. The resulting force field simulates a dynamic system that seeks equilibrium—class-specific centers around which vertex embeddings self-organize.

In the next section, we formalize this force-based mechanism within our proposed model, GRAVITY.

\section{Methodology}\label{sec: methodology}
\subsection{Attractive Force}

In classical physics, gravitational mass acts as a global scalar that determines how strongly an object can attract or be attracted by another. In graph structures, however, vertices are embedded within a topology that imposes local connectivity constraints through edges. Thus, the influence of a vertex \( o_i \) on another vertex \( o_j \), denoted by the attractive force \( f_{i,j} \), is mediated by both structural proximity and semantic alignment. Importantly, this influence is generally \emph{not symmetric}, even in undirected graphs, due to unequal neighborhood densities or edge weights. Moreover, the impact one vertex may have on another can be modified whenever the structural shape of the graph changes. This means that the notion of ``mass'' in the graph context intrinsically depends on the topology and distribution of local influences.

Considering the object attributes, we define the attractive force as
\begin{align}\label{Eq:node_2_node_force}
f_{\lambda}(o_{i}, o_{j}\,|\,x_i, x_j, \mathcal{G}) &= \nabla \Phi(o_i,o_j|\mathcal{G}, \lambda) \\
    &= \begin{cases}
        s_{i,j} \times  t_{i,j} \times \mathbf{W}_{\mathcal{E}} \;\;\; \mbox{    if } \text{Path}_{i}^{j} \neq \emptyset \\ 
        \;\;\;\; \text{ and } s_{i,j} \times  t_{i,j} \geq \lambda\,, \\
        0 \mbox{ otherwise}
    \end{cases} \notag\, ,
\end{align}
where \(\text{Path}_i^j\) is the shortest path from vertex \( o_i \) to vertex \( o_j \) in \( \mathcal{G} \). The term \( s_{i,j} = \mathit{sim}(x_i, x_j) \) denotes a similarity score measuring attribute-level resemblance between \( x_i \) and \( x_j \). The term \( t_{i,j} \) represents the \emph{social tie}—a directional, structure-aware factor modulating how much influence flows from vertex \( o_j \) to vertex \( o_i \), while \( \Theta_{\mathcal{E}} \) encodes a learnable intensity modulation reflecting spatial or positional characteristics. The threshold \( \lambda \) filters out negligible interactions.

The social tie \( t_{i,j} \in [0,1] \) quantifies the effective structural influence from \( o_j \) to \( o_i \), increasing with strong local cohesion and decreasing over longer or weaker paths. It can be explicitly modeled using a normalized product of edge-neighborhood contributions along the path from \( o_j \) to \( o_i \), as follows:
\begin{align}\label{Eq:social ties}
    t_{i,j} &= \prod_{\substack{(o_{\alpha}, o_{\beta}) \\ \in \; \text{Path}_{i}^j}} 
    \frac{
        \sum\limits_{o_{\tau} \in \bm{\mathcal{N}}_{\alpha}} \omega_{\alpha,\tau}
    }{
        \max\left\{ 
            \sum\limits_{o_{\tau} \in \bm{\mathcal{N}}_{\alpha}} \omega_{\alpha,\tau}\;,\;
            \sum\limits_{o_{\tau} \in \bm{\mathcal{N}}_{\beta}} \omega_{\beta,\tau} 
        \right\}
    }\,,
\end{align}
where \( \omega_{\alpha,\tau} \) represents the interaction weight from node \( o_{\alpha} \) to its neighbor \( o_{\tau} \), and \( \bm{\mathcal{N}}_{\alpha} \) denotes the neighborhood of node \( o_{\alpha} \).

However, this formulation is not only computationally expensive for large graphs, but also \emph{transductive}, requiring complete knowledge of the graph to evaluate \( t_{i,j} \). To enable scalable and \emph{inductive} computation of these social ties; especially in evolving or partially observed graphs, we propose to approximate the social ties using a graph neural function:
\begin{align}\label{Eq:social_tie_gnn}
    t_{i,j} &\approx \text{sigmoid}(\text{GNN}(o_i,o_j|\mathcal{G}))
\end{align}
where the GNN is trained to approximate social ties (as presented in Eq.~(\ref{Eq:social ties})) from the graph.

Consequently, attractive forces between vertices are modulated by these learned social ties, with higher values inducing stronger pulls. The entire interaction pattern across the graph is shaped by the distribution of these mass-like tie coefficients. For a set of objects \( \bm{\mathcal{O}} \), the matrix of all pairwise attractive interactions is given by \( \mathbf{T} = (t_{i,j}) \).

\begin{align}\label{Eq:whole-node-2-node-force}
    \mathbf{F}_{\lambda} &= (\mathbf{S} \odot \mathbf{T})\cdot \Theta_{\mathcal{E}} \,,
\end{align}
where \( \mathbf{S} = (s_{i,j}) \) is the similarity matrix and \( \odot \) denotes element-wise multiplication.

Finally, to compute the influence of groups on individual vertices, we define the group-to-vertex force matrix \( \mathbf{F}_{g, \lambda} \) using a binary membership matrix \( \mathbf{M} \in \{0,1\}^{N \times K} \), where each entry indicates group affiliation:
\begin{align}\label{Eq:group_2_node_force}
    \mathbf{F}_{g, \lambda} &= \mathbf{F}_{\lambda} \cdot \mathbf{M}\,.
\end{align}

\begin{figure*}[!t]
    \centering
    \includegraphics[width=.85\linewidth]{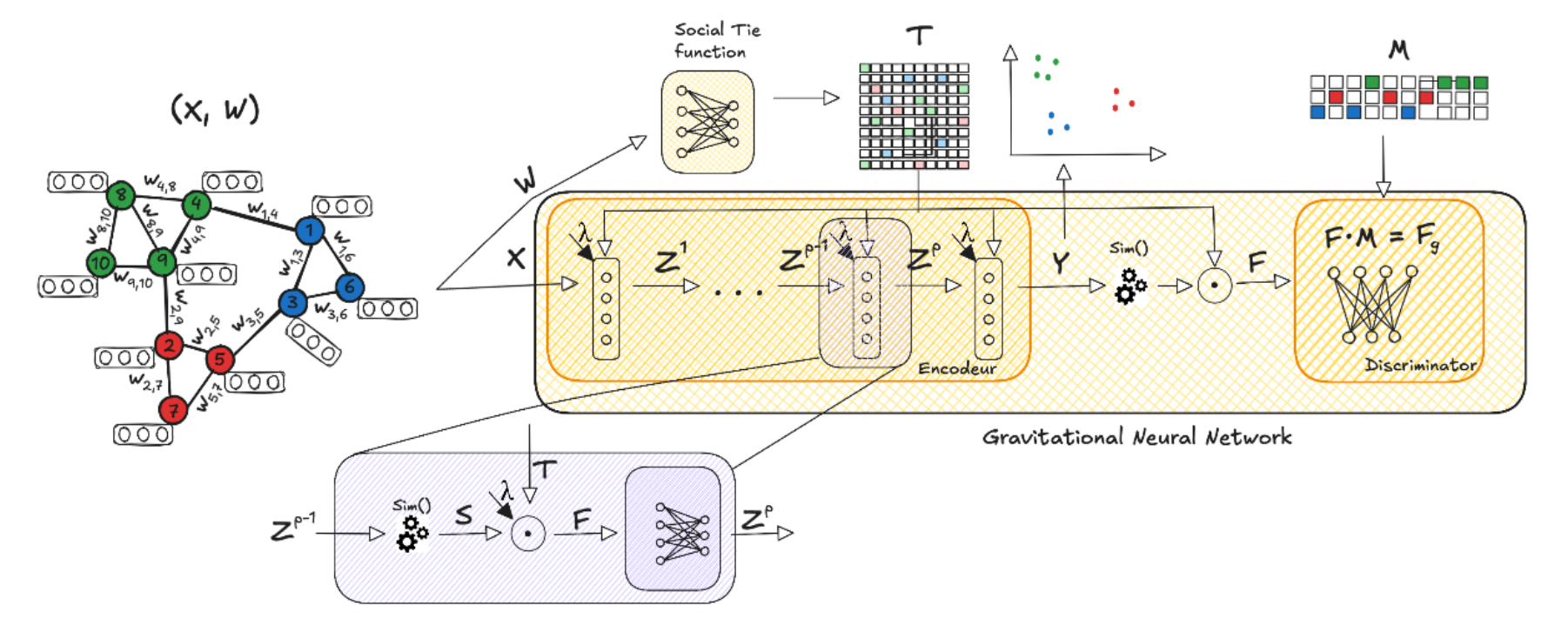}
    \caption{GRAVITY Framework}
    \label{fig:gravity-overview}
\end{figure*}

\subsection{Gravitational neural network}
Based on the attractive force definition in Eq.~(\ref{Eq:node_2_node_force}), the underlying gravitational \textit{potential function} \( \Phi \) that governs the movement of objects (and relates their position) in the latent field can be formalized as:
\begin{align}
    \Phi(o_i|\mathcal{G},\,\lambda) &= y_i \\
    &= \int_{\substack{o_{j}\in \; \bm{\mathcal{O}} \\ o_j \neq o_i }} f_{\lambda}(o_i, o_j|x_i, x_{j},\mathcal{G}) \, dx_j \notag\,, 
\end{align}
defines the potential landscape in which vertex \( o_i \) evolves. This formulation reflects the classical physical principle where entities move along gradients toward lower potential energy, driven here by the interplay of semantic alignment and topological closeness.

To simulate this gravitational evolution and compute the object’s position in the latent field, we define an encoder with parameters $\Theta_{\mathcal{E}} = \{\mathbf{W}_{\mathcal{E}}^p, b_{\mathcal{E}}^p\}_{p=1}^P$ given as:
\begin{align}\label{Eq:encoder-net-1}
    \Phi(x_i|\bm{\mathcal{G}}, \Theta_{\mathcal{E}}, \lambda) &= \code{x_i}{G}{E} \\
    &= \sigma\Bigl(\mathbf{W}^P_{\mathcal{E}}\cdot z_i^{P} + b^{P}_{\mathcal{E}}\Bigr),\notag 
\end{align}
where $\mathbf{W}_{\mathcal{E}} =  \{\mathbf{W}_{\mathcal{E}}^p\}_{p=1}^P$, \( z_i^P \) a recursively aggregated latent signal that reflects the cumulative influence from other vertices. Specifically, it evolves layer by layer through gravitational aggregation:
\begin{align}\label{Eq:intermediary_layer_encoder-1}
    z_i^p &= \phi\Biggl(\mathbf{W}^{p-1}_{\mathcal{E}} \cdot \sum_{\substack{o_{j}\in \; \bm{\mathcal{O}} \\ o_j \neq o_i }}
    f_{\lambda}(o_i, o_j|z_i, z_j,\mathcal{G}) \Biggr),
\end{align}
In this view, the encoder is guiding each vertex to its equilibrium position in a gravitationally organized latent space.

To assess whether the latent representations reflect coherent gravitational behavior, we reevaluate the induced force field between vertices ($\mathbf{F}^{'}_{\lambda}$) and substructures ($\mathbf{F}^{'}_{g,\lambda}$), using definitions from Equations~(\ref{Eq:whole-node-2-node-force}) and~(\ref{Eq:group_2_node_force}). These matrices serve as inputs to the discriminator, which evaluates whether the gravitational coherence of each vertex with respect to candidate substructures holds.

The discriminator is also a $R$-layer neural network with parameters $\Theta_{\mathcal{D}} = \{\mathbf{W}^r_{\mathcal{D}},\,b^r_{\mathcal{D}}\}_{r=1}^R$, designed to interpret each row $\mathbf{F}^{'}_{g,\lambda,i}$ of $\mathbf{F}^{'}_{g,\lambda}$ as a gravitational signature vector. The predicted group affinity of vertex $o_i$ is given by:
\begin{align}\label{Eq:discriminator}
    \decode{{F}^{'}_{g,\lambda,i}}{D} &= \mbox{\it Softmax}\left(\mathbf{W}_{\mathcal{D}}^R \cdot h^R_i  + b^R_{\mathcal{D}}\right),
\end{align}
where intermediate layers are updated recursively via:
\begin{align}\label{Eq:first_layer_discriminator}
    h^r_i &= \phi\left(\mathbf{W}^{r-1}_{\mathcal{D}} \cdot h^{r-1}_i + b^{r-1}_{\mathcal{D}} \right),
\end{align}
with $h^0_i = \mathbf{F}^{'}_{g,\lambda,i}$. Through this process, the gravitational neural network learns to organize the latent space such that attractive forces reflect the graph’s intrinsic structure, encouraging the emergence of well-separated, semantically meaningful substructures. In Figure~\ref{fig:gravity-overview}, we have an overview of the GRAVITY framework, which encodes vertex features and relational weights into a latent space through a gravitational neural network. The resulting force field \(\mathbf{F}\) of the gravitational network guides downstream discrimination through a decoder.

\subsection{Learning process}
The representation learning goal is to discover an energy-efficient layout in the latent space, where intra-group cohesion and inter-group repulsion are in balance, resembling gravitational equilibrium.

This approach extends principles from force-directed layouts \cite{fruchterman1991graph,kobourov2012spring} and adaptive partitioning schemes \cite{ye2008adaptive}, repurposing them to guide the learning dynamics of a neural model. At each iteration, the encoder adjusts vertex representations to reflect the intensity of virtual gravitational pull induced by other vertices, based on structural proximity and attribute alignment. The encoder’s objective is thus to minimize internal energy: placing each vertex closer to its gravitational center (i.e., its group) while avoiding collapse into unrelated groups.

To achieve this, we define a force-aligned loss function. Let \( \bm{\mathcal{Y}} = \bigcup_{\kappa=1}^K \bm{\mathcal{Y}}_{\kappa} \) denote the latent embeddings. We expect embeddings within the same group to be similar and dissimilar to those of other groups. This dynamic is quantified through a silhouette-based encoder loss:

\begin{align}\label{Eq:encoder-loss}
    \losscode{E} &= \underset{\Theta_{\mathcal{E}}}{\min}\sum_{i=1}^N \frac{1}{2}\left(1 - \silh{y_i}\right),
\end{align}
where the silhouette score \(\silh{y_i}\) measures how well each vertex \(o_i\) is attracted toward its assigned group while repelling others:
\begin{align}
    \silh{y_i} &= \frac{\In{y_i}- \Out{y_i}}{\max\left\{\In{y_i},\, \Out{y_i}\right\}},\\
    \In{y_i} &= \sum_{\kappa=1}^K \delta_{i,\kappa} \sum_{\substack{y_j\in \bm{\mathcal{Y}}_{\kappa}\\ j\neq i}} \mathit{sim}(y_{i}, y_{j}), \\
    \Out{y_i} &= \min_{\kappa' \neq \kappa} \sum_{y_j\in \bm{\mathcal{Y}}_{\kappa'}}\mathit{sim}(y_{i}, y_{j}),
\end{align}
where \( \delta_{i,\kappa} \) indicates group membership.

After each forward pass, we recompute the force field generated by the updated embeddings using Equations~\eqref{Eq:whole-node-2-node-force} and~\eqref{Eq:group_2_node_force}, yielding the force matrices $\mathbf{F}^{'}_{\lambda}$ and $\mathbf{F}^{'}_{g,\lambda}$. In an ideal gravitational scenario, vertex $o_i$ should experience maximal attraction to its native subgraph $g_{\kappa}$ while remaining minimally affected by other subgraphs. We represent this gravitational ideal through the target vector $\overline{\mathbf{F}}_{g_{\kappa},\lambda, i}$, where only the $\kappa^{th}$ dimension contains positive values while all others are zero.

To enforce this behavior, the discriminator learns to classify gravitational alignment by evaluating whether each force vector \( \mathbf{F}^{'}_{g,\lambda,i} \) adheres to the desired gravitational attraction profile:
\begin{align}
    \lossdiscr{D} = \underset{\Theta_{\mathcal{D}}}{\min} \sum_{i=1}^N \left( 1 - \decode{{F}^{'}_{g,\lambda,i}}{D} \right).
\end{align}

The final training objective becomes:
\[
\mathcal{L}\mathit{oss}(\Theta_{\mathcal{E}},\, \Theta_{\mathcal{D}}) = \losscode{E} + \gamma \, \lossdiscr{D},
\]
where \( \gamma > 0 \) regulates the influence of the discriminator’s gravitational consistency check.

\begin{figure}[!t]
    \centering
    \subfloat[Cora epoch 1]{\includegraphics[width=.162\textwidth]{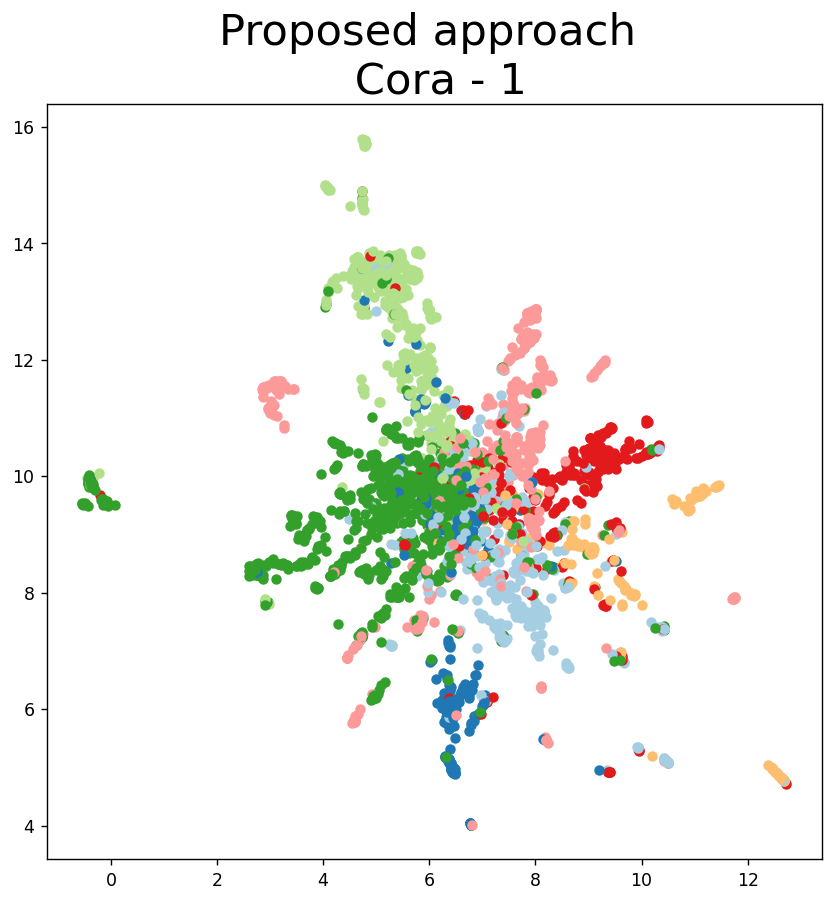}}
    \subfloat[Cora epoch 5]{\includegraphics[width=.162\textwidth]{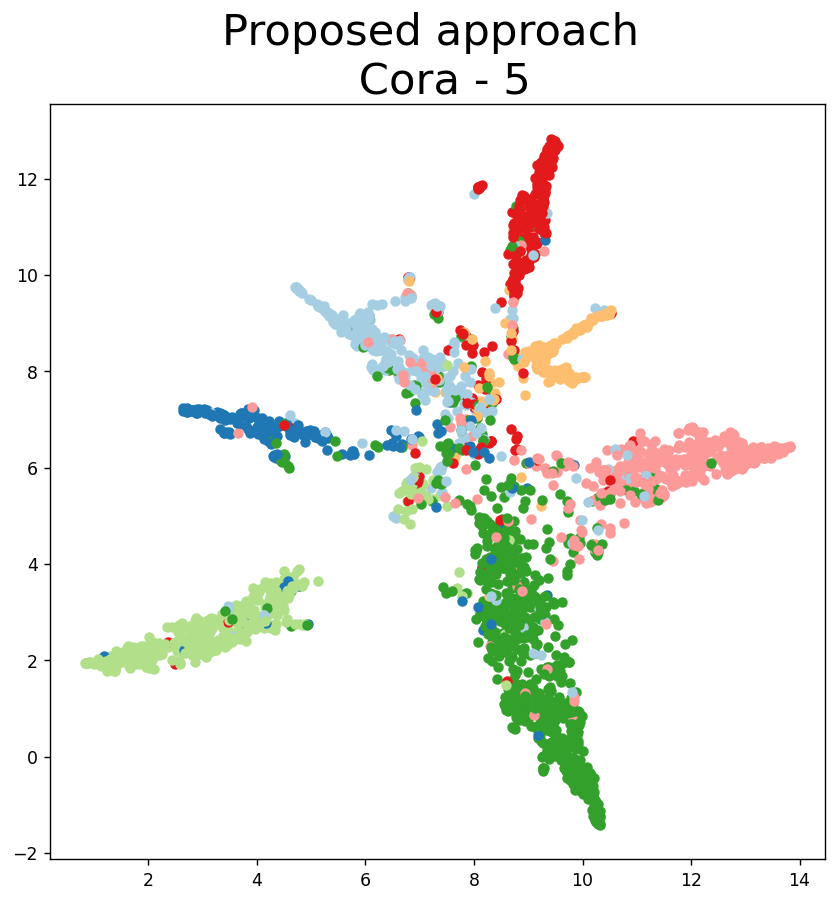}}
    \subfloat[Cora epoch 200]{\includegraphics[width=.162\textwidth]{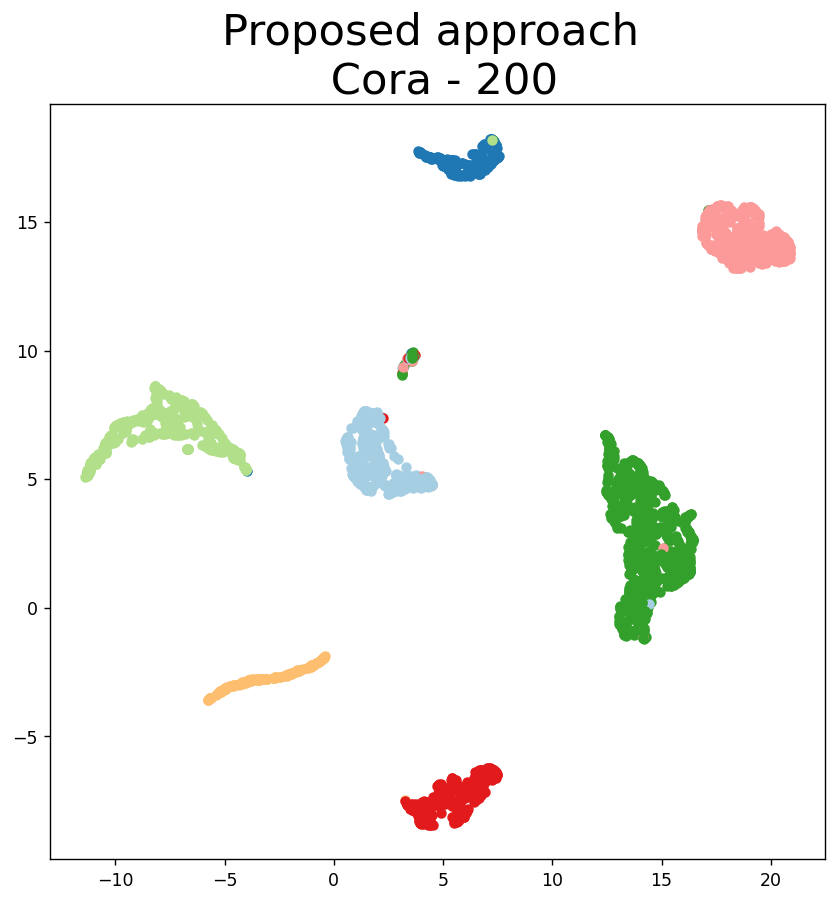}}
    \caption{Embedding of the vertices at different epochs in Cora data.}
    \label{fig:node-embedding-visualization-Cora}
\end{figure} 

\begin{figure*}
    \centering
    \begin{tabular}{ccc}
    In-domain evaluation & \multicolumn{2}{c}{Zero-shot evaluation}\\
        \includegraphics[width=.315\linewidth]{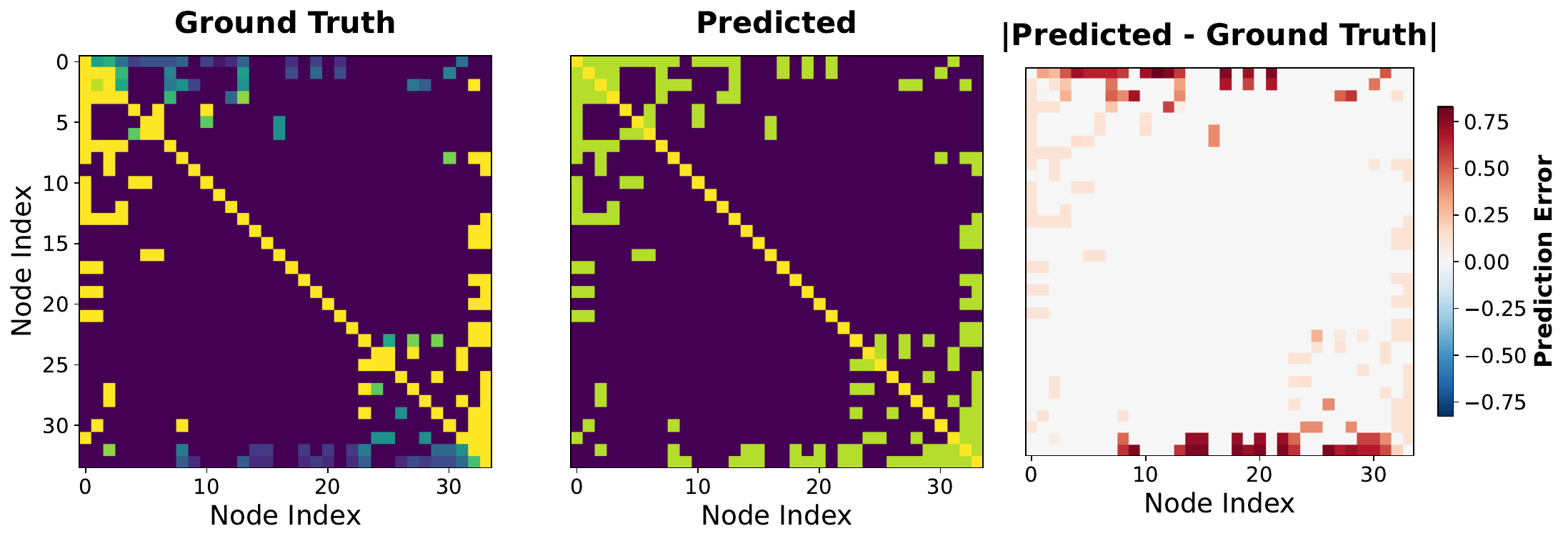} & \includegraphics[width=.315\linewidth]{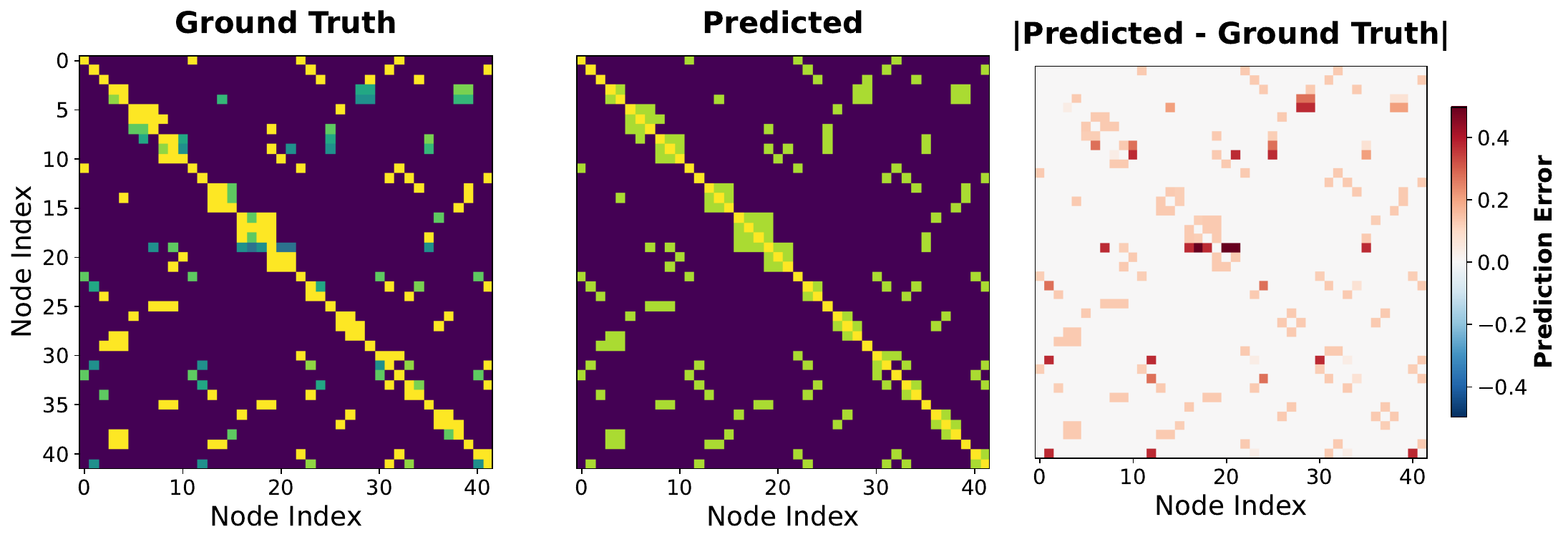} & \includegraphics[width=.315\linewidth]{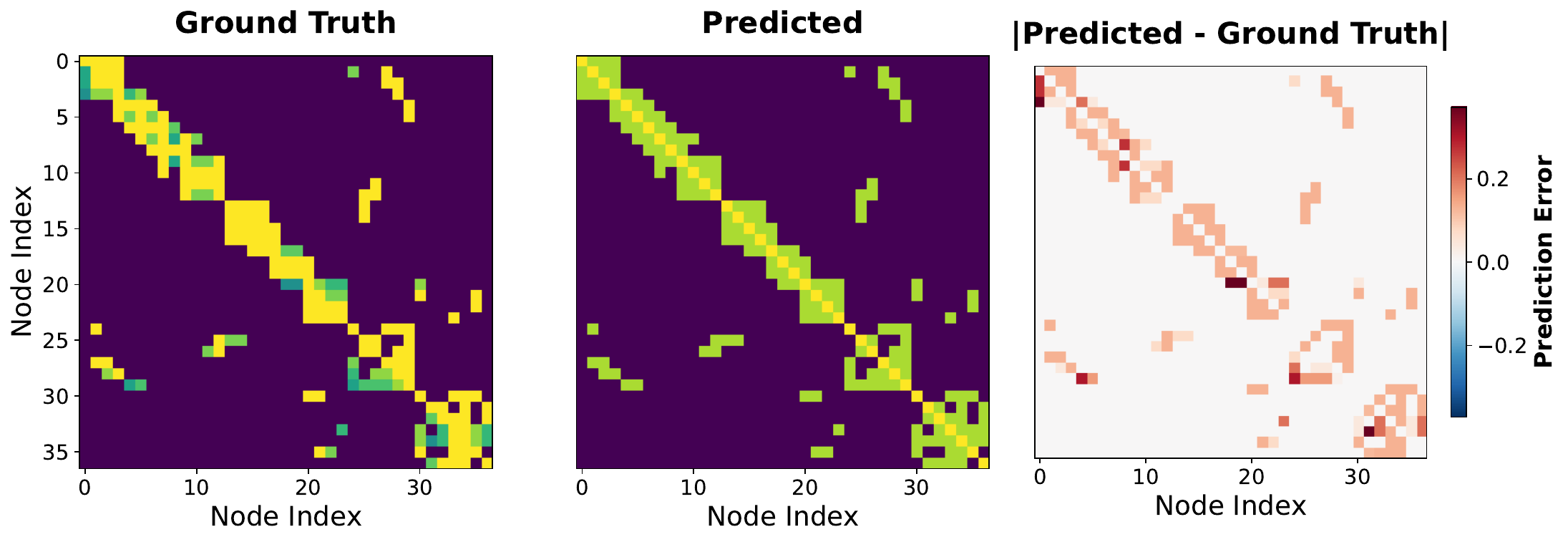} \\
         KARATE & PROTEINS & ENZYMES 
    \end{tabular}
    \caption{Social tie model performance.}
    \label{fig:model generalization}
\end{figure*}
As shown in Fig.~\ref{fig:node-embedding-visualization-Cora}, our field-driven training process iteratively organizes the embedding space into distinct orbital structures, forming coherent group-specific attractors similar to gravitational basins. Here, vertices naturally cluster according to their subgraph affiliations.

In summary, GRAVITY exhibits natural inductive behavior by design, owing to its learned force-driven mechanism that generalizes across heterogeneous graph components. Rather than memorizing node-specific representations, the model learns a continuous mapping function that governs the aggregation and interaction of vertices based on structural proximity and attribute-level similarity. This allows GRAVITY to accommodate unseen vertices or subgraphs without retraining, dynamically recalibrating the latent gravitational field to preserve semantic coherence and class-aligned attractors. The adaptive receptive fields guided by force dynamics further enable context-aware integration, ensuring that new components contribute meaningfully to the embedding space while maintaining geometric and class-level organization. 

\section{Experiments}
\label{sec:experiments}

\begin{table}[!t]
    \centering
    \resizebox{.45\textwidth}{!}{\begin{tabular}{|l|lll|} 
    \multicolumn{4}{c}{In-domain evaluation}\\
    \hline
       Datasets  &  MAE & MSE & MAPE \\ \hline
       Karate & $0.16\pm 0.01$ & $0.03 \pm 0.002$ & $0.02 \pm 0.001$ \\
       MUTAG & $0.14 \pm 0.02$ & $0.024 \pm 0.01$ & $0.013 \pm 0.001$ \\
       ENZYMES & $0.11 \pm 0.003$ & $0.02 \pm 0.001$ & $0.011 \pm 0.0003$ \\
       PROTEINS & $0.14 \pm 0.005$ & $0.03 \pm 0.0004$ & $0.013 \pm 0.001$ \\
       NC11 & $0.15 \pm 0.013$ & $0.03 \pm 0.002$ & $0.015 \pm 0.001$ \\
       \hline
       \multicolumn{4}{c}{Zero-shot evaluation}\\ \hline
       Datasets  &  MAE & MSE & MAPE \\ \hline
       MUTAG & $0.003 \downarrow$ & $0.03\uparrow$ & $0.68\uparrow$ \\
       ENZYMES & $0.001\downarrow$ & $0.02\downarrow$ & $0.54\uparrow$ \\
       PROTEINS & $0.001\downarrow$ & $0.014\downarrow$ & $0.51\uparrow$ \\
       NC11 & $0.004\downarrow$ & $0.03$ & $0.51\uparrow$ \\
       \hline
    \end{tabular}}
    \caption{Generalization error of the social tie model. The model is trained on the Karate dataset and applied to generate social ties in the MUTAG, ENZYMES, PROTEINS, and NCI1 datasets. $\uparrow$ indicates a decrease in performance, while $\downarrow$ denotes an improvement.}
    \label{tab:model generalization}
\end{table}
\subsection{Experimental setup}
\paragraph{Datasets} We evaluate our method on widely-used node classification benchmarks, including Cora, Citeseer, Pubmed, and Amazon-Photo. To assess the effectiveness of the proposed social tie modeling, we further conduct experiments on well-known graph datasets such as Karate, MUTAG, ENZYMES, PROTEINS, and NCI1.

\paragraph{Experimental Protocol}
We adopt the standard transductive protocol for Cora, Citeseer, and Pubmed as in~\cite{kipf2017semi}, and report average accuracy over multiple random splits for Chameleon and Squirrel. Models are trained using the Adam optimizer (learning rate 0.01, weight decay \(5 \times 10^{-4}\)) with early stopping based on validation accuracy (patience of 100 epochs). Key hyperparameters of \textsc{GRAVITY}, including the latent space dimension, gravitational strength \(\lambda\), and learning rate, are tuned using the validation set. All experiments are implemented with PyTorch Geometric and executed on an MPS backend.

Our evaluation is organized in three parts. First, we examine the ability of \textsc{GRAVITY} to infer social ties (Eq.~\ref{Eq:social ties}) under both few-shot and zero-shot conditions, where generalization is tested across structurally different graphs. Second, we visualize the node embeddings on Citeseer, Cora, and Pubmed to illustrate how gravitational interactions shape representation topology across multiple hops. Lastly, we evaluate vertex classification performance in comparison to recent approaches.

\paragraph{Baselines}
We benchmark against 6 GNN models including GraphSAGE~\cite{hamilton2017inductive}, GAT~\cite{velivckovic2018graph}, RNCGLN~\cite{zhu2024robust}, GRACE~\cite{zhu2021contrastive},  BIAS~\cite{zhao2021graph}, GRLC~\cite{peng2023grlc} and DGI~\cite{velivckovic2018deep}.

\begin{figure*}[]
    \centering
    \begin{tabular}{cccc}
       & Train sample  & \cellcolor{black!25} & Test sample \\
       \multirow{-5}{*}{\rotatebox[origin=c]{90}{Cora}} & \includegraphics[width=.5\linewidth]{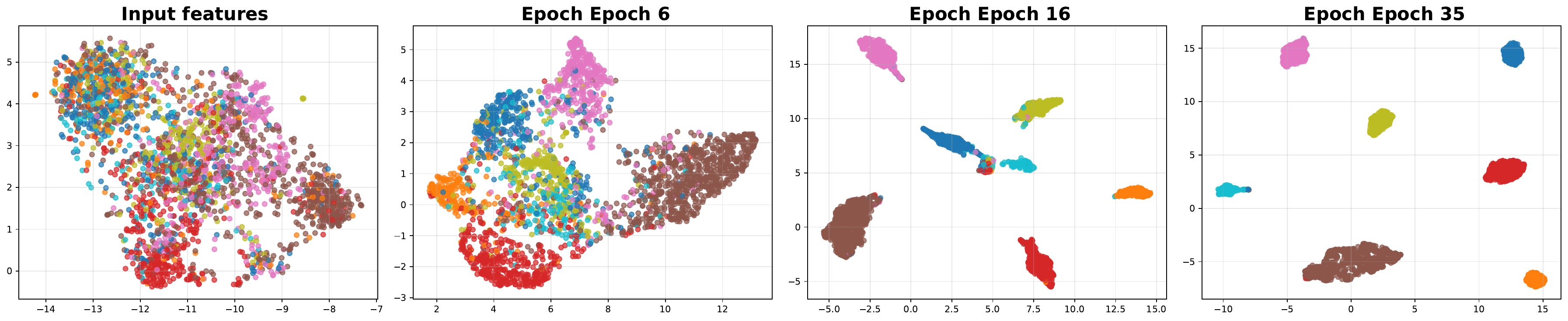}  & \cellcolor{black!25} & \includegraphics[width=.25\linewidth]{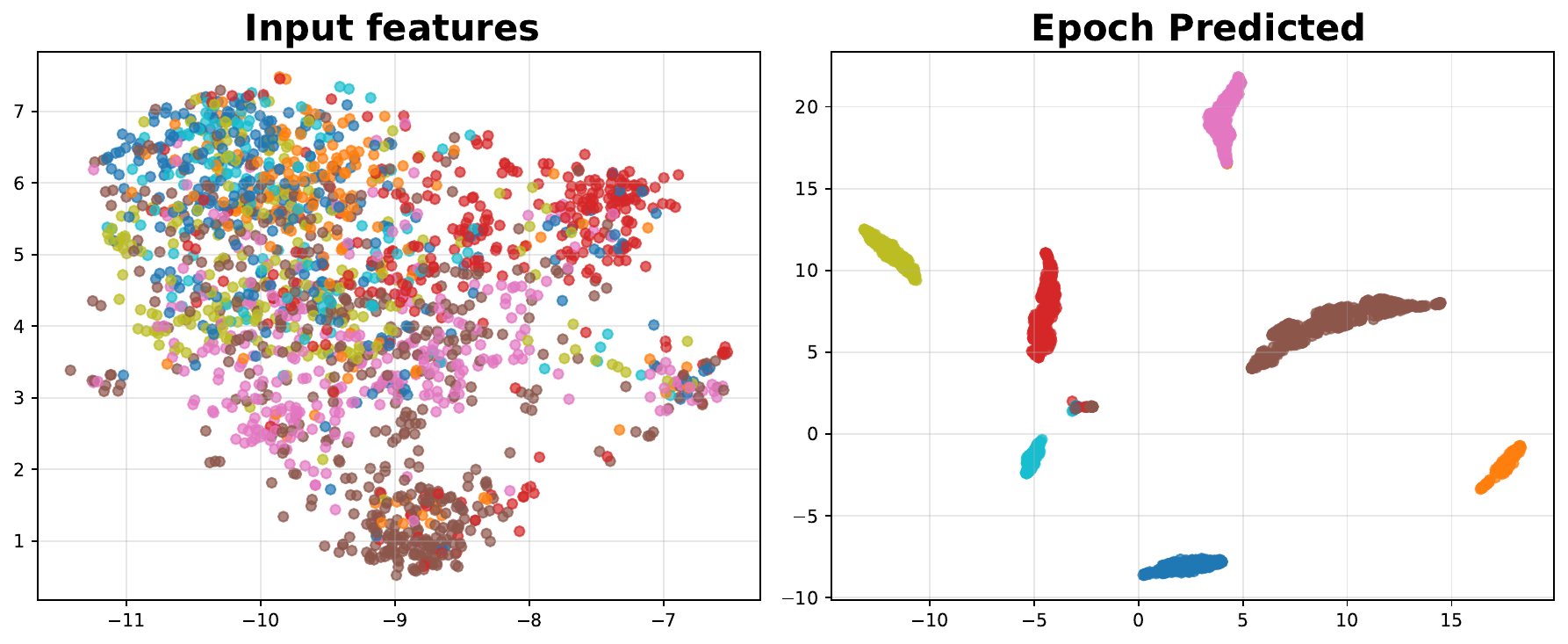} \\ 
       \multirow{-5}{*}{\rotatebox[origin=c]{90}{Citeseer}} & \includegraphics[width=.5\linewidth]{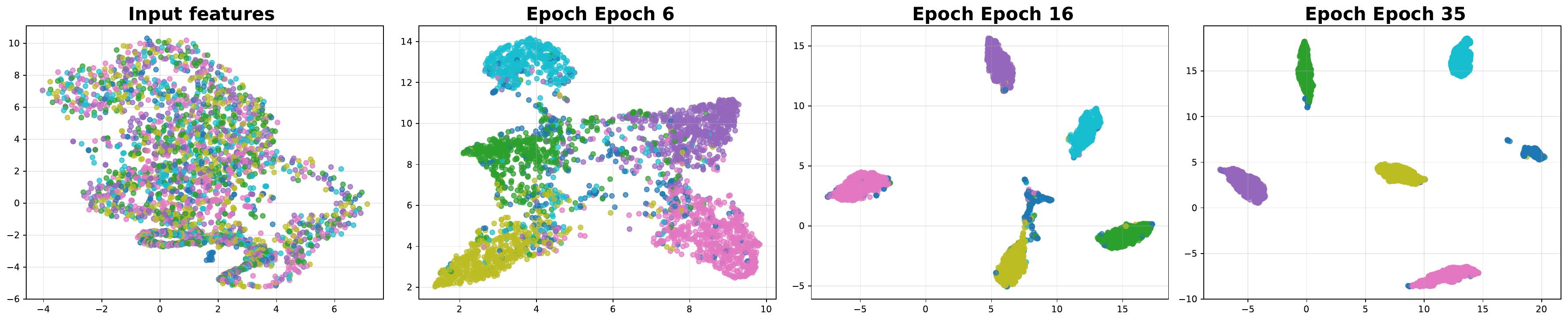}  & \cellcolor{black!25} & \includegraphics[width=.25\linewidth]{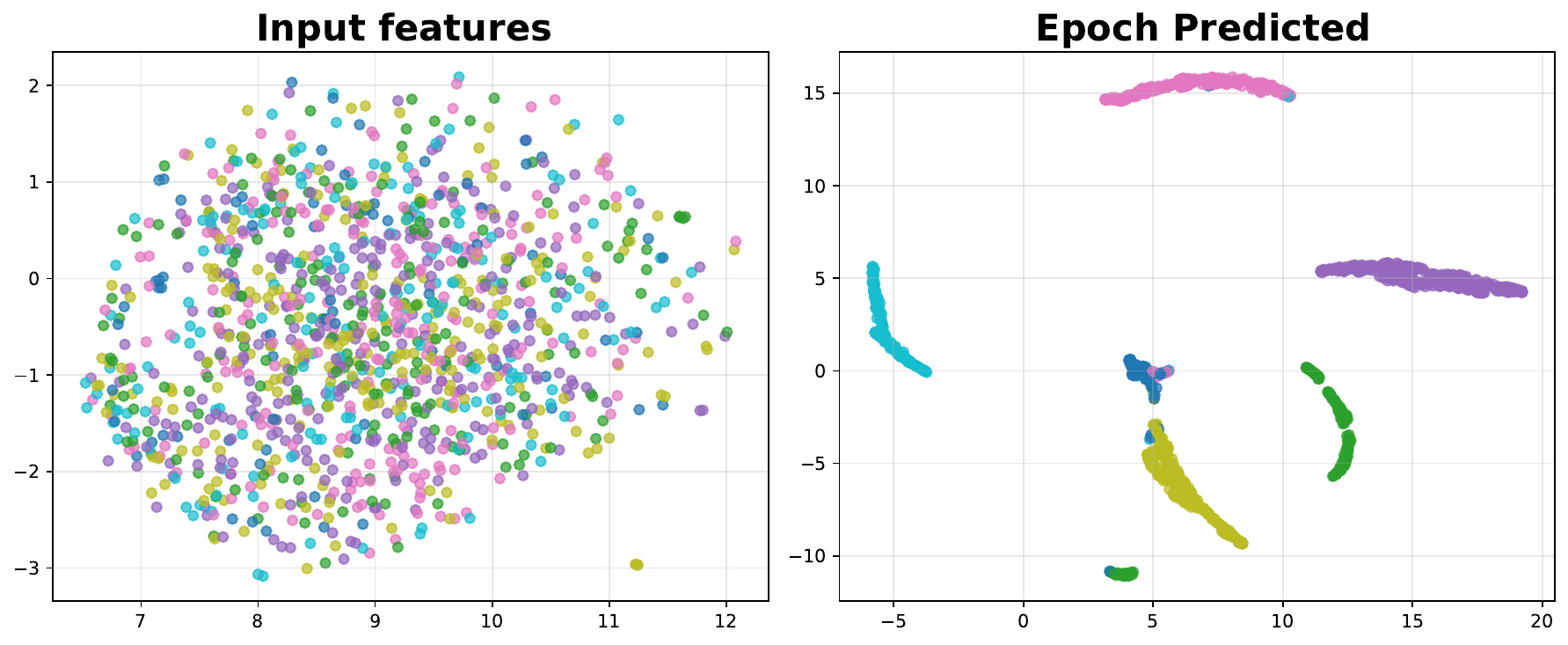} \\ 
       \multirow{-5}{*}{\rotatebox[origin=c]{90}{Amazon}} & \includegraphics[width=.5\linewidth]{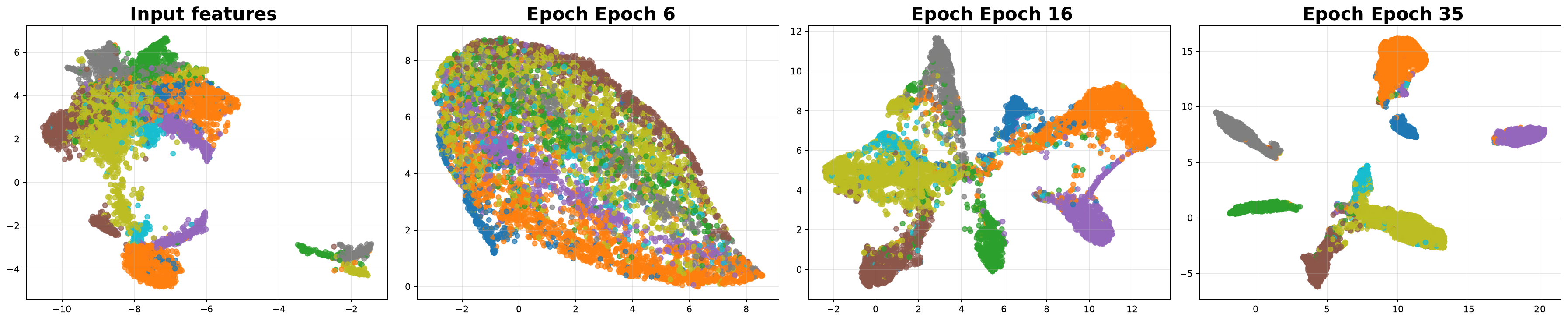}  & \cellcolor{black!25} & \includegraphics[width=.25\linewidth]{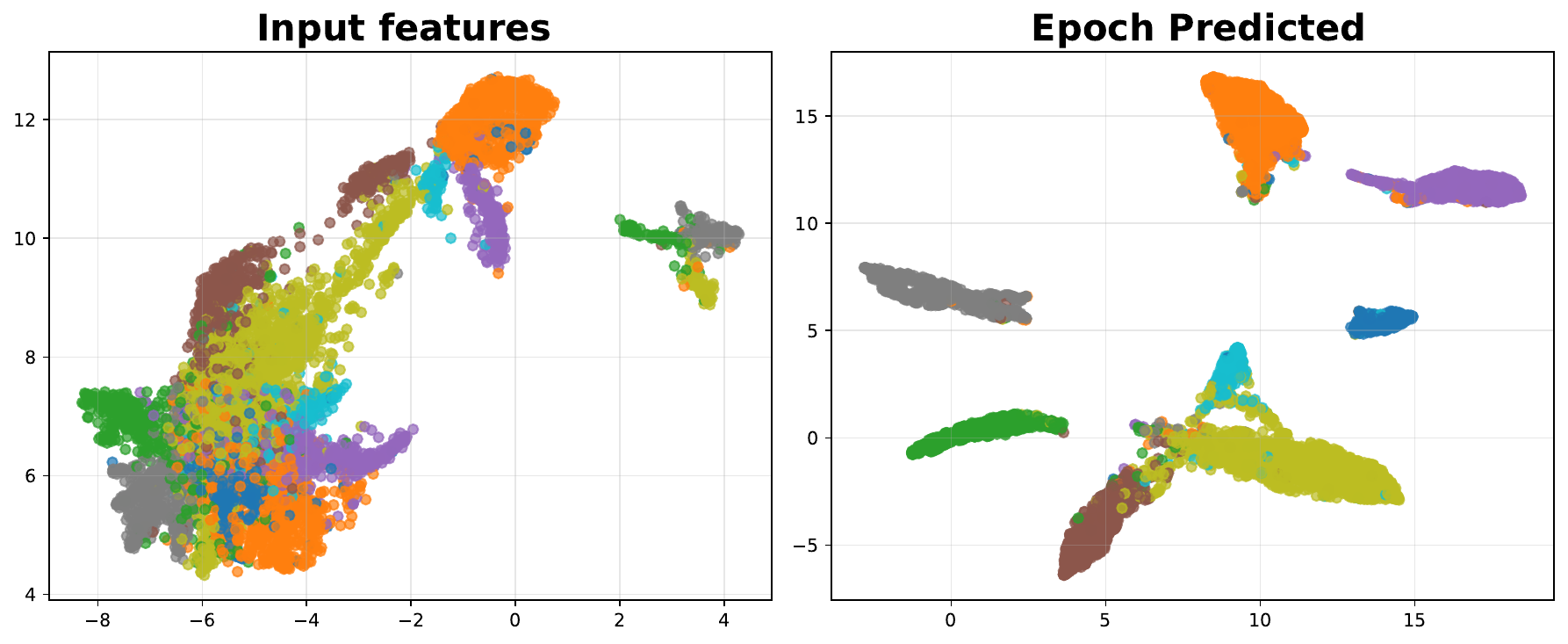} \\
    \end{tabular}
    \caption{GRAVITY embeddings on Cora, Citeseer and Amazon (Photos) graphs under an inductive setup.}
    \label{fig:train_test_umap}
\end{figure*}

\subsection{Social ties}
The in-domain results on the Karate dataset indicate that the social tie model accurately captures structural interactions, exhibiting low mean absolute error (MAE = 0.16), mean squared error (MSE = 0.03), and mean absolute percentage error (MAPE = 0.02). When extended to additional datasets such as MUTAG, ENZYMES, PROTEINS, and NCI1, the model maintains strong predictive capacity with comparable in-domain errors. Particularly, ENZYMES and PROTEINS report the lowest MAE and MAPE, suggesting robust modeling of inter-vertex relations. As illustrated in Figure~\ref{fig:model generalization}, the predicted tie matrices visually match the ground truth, with error matrices showing sparse and localized deviations.

In the zero-shot evaluation setting, where the model is trained on Karate but deployed on unseen datasets, we observe a nuanced behavior. While the MAE and MSE either remain stable or even decrease (as denoted by $\downarrow$), MAPE increases markedly across all target datasets. This indicates that although absolute prediction errors are low, the model’s relative error becomes more pronounced—likely due to smaller ground truth values amplifying the percentage difference. Despite this, the overall generalization performance demonstrates the model’s potential in transferring learned tie dynamics across different graph domains.

\subsection{Embeddings}
Figure~\ref{fig:train_test_umap} demonstrates how GRAVITY evolves vertex embeddings over training epochs on both Citeseer and Amazon datasets, under an inductive setup where training and test graphs are disjoint. Initially dispersed in input space, the vertices progressively converge into compact, class-cohesive clusters as training proceeds. This organization is driven by the gravitational field induced by GRAVITY’s learned social tie model, where each class acts as a latent attractor pulling similar vertices together.

Crucially, this gravitational field extends to unseen graphs. Even though test graphs were not observed during training, their vertex embeddings align around pre-established attractors, allowing GRAVITY to naturally impose structure on new data. Novel vertices are drawn into the latent force field and smoothly incorporated into their inherent class groups, reflecting GRAVITY’s strong generalization ability in scenarios where graph topology is not directly transferred.

\begin{table}[!t]
    \centering
    \caption{Node classification accuracy (\%, mean ± std) under the transductive setting on four benchmark datasets. GRAVITY consistently outperforms other methods across Cora, Citeseer, and Amazon-Photo, demonstrating strong representational power. The GRAVITY model is tested under inductive setting.}
    \resizebox{.45\textwidth}{!}{\begin{tabular}{l|llll} \multicolumn{5}{c}{Transductive setting}\\ \hline
       Methods  & Cora & Citeseer & Pubmed & Amazon-Photo\\ \hline
        SAGE & $86.90\pm 1.04$ & $76.04\pm 1.30$ & $88.45\pm 0.50$ &  $91.6 \pm 0.6$ \\ 
        GAT & $83.0 \pm 0.7$ & $72.05 \pm 0.7$ & $79.01 \pm 0.3$& $91.8 \pm 0.6$\\
        RNCGLN & $84.46\pm 0.35$ & $73.96\pm 0.30$ & $81.16\pm0.64$ & $92.21\pm 0.54$  \\ 
        GRACE & $83.2\pm 0.4$ & $72.1\pm 0.5$ & $86.01 \pm 0.1$ & $92.78 \pm 0.45$  \\ 
        BIAS & $87.8\pm 0.5$ & $74.5\pm 0.6$ & $81 \pm 0.5$ & $82.38 \pm 0.45$  \\
        GRLC (3-hops) & $83.5\pm0.5$ & $72.6 \pm 0.6$ & $82.1 \pm 0.4$ &  $92.3\pm 0.5$ \\ 
        DGI & $82.6\pm0.4$ & $68.8\pm0.7$ & $86.0\pm0.1$ &  $86.7 \pm 0.37$ \\  \hline
        \multicolumn{5}{c}{Inductive setting}\\ \hline
        GRAVITY &  $\bm{92.28 \pm 0.23}$ & $\bm{92.11 \pm 0.32}$ & $\bm{90.45 \pm 0.6}$ & $\bm{94.51\pm 0.21}$\\ \hline
    \end{tabular}}
    \label{tab:comparative study}
\end{table}

\subsection{Classification tasks}
Table~\ref{tab:comparative study} presents a comparative evaluation of GRAVITY against state-of-the-art baselines across four benchmark datasets: Cora, Citeseer, Pubmed, and Amazon-Photo. While all competing methods are evaluated under the transductive setting, GRAVITY is tested under the more challenging inductive setting where the test graph remains unseen during training.

Despite this discrepancy, GRAVITY achieves superior performance across all datasets, with the highest accuracy on Cora ($92.28\%$), Citeseer ($92.11\%$), Pubmed ($90.45\%$), and Amazon-Photo ($94.51\%$). These results highlight the model’s strong generalization capacity and its ability to leverage the gravitational interaction mechanism to dynamically guide vertex representation, even in unseen graph structures. This inductive strength positions GRAVITY as a robust and scalable alternative to classical message-passing or contrastive approaches.

\section{Related work}

Dynamic graph neural networks incorporate temporal modeling to capture evolving graph structures \cite{zheng2025survey}, but they often rely on fixed aggregation rules. Recent methods add adaptivity: for example, SEAN \cite{zhang2024towards} selects representative neighbors and decays outdated information; RTR \cite{chen2023recurrent} uses node-wise RNNs to accumulate all past neighbors; Ada-DyGNN \cite{li2024robust} employs reinforcement learning to choose which neighbors to update; and ASGNN \cite{zhang2022asgnn} learns to refine the adjacency matrix during message passing. Foundational inductive frameworks like GraphSAGE \cite{hamilton2017inductive} learn node representations via trainable neighborhood aggregation functions, enabling generalization to unseen nodes but relying on fixed sampling schemes. GATs \cite{velivckovic2018graph} improve adaptivity further via masked self-attention layers, allowing nodes to implicitly assign different weights to neighbors without upfront graph knowledge and enabling inductive generalization. These approaches adapt the structural context through dynamic neighbor selection or weighting. Likewise, community- and motif-aware models explicitly enforce higher-order structure: CLEAR \cite{zhu2023unsupervised} and DMoN \cite{tsitsulin2023graph} incorporate clustering objectives, while motif-based methods such as the Neural Graph Pattern Machine \cite{wang2025beyond} and ASE-Mol \cite{jiang2025adaptive} sample and encode task-relevant substructures. In contrast, GRAVITY’s novelty is an implicit, physics-inspired mechanism: it learns a gravitational field over the graph where structurally and semantically related nodes continuously attract. This latent force-based grouping warps the embedding space during training so that nodes in the same community or sharing motifs naturally draw together, without requiring separate clustering or motif modules.

Self-supervised GNNs have also sought to align structure and features without labels. Contrastive methods like GraphCL \cite{you2020graph} create multiple augmented views, while generative models like GraphMAE \cite{hou2022graphmae} reconstruct masked node features. Techniques such as SimGRACE \cite{xia2022simgrace} and CCA-SSG \cite{zhang2021canonical} avoid complex augmentations by perturbing the encoder or decorrelating features, and DGPM \cite{yan2024empowering} jointly reconstructs nodes and discovers salient subgraph motifs. In the same vein, GRACE \cite{zhu2021contrastive} optimizes node-level agreement by contrasting embeddings from dual corrupted graph views (via edge removal and feature masking), theoretically linking to mutual information maximization. Similarly, DGI \cite{velivckovic2018deep} maximizes mutual information between local patch representations and a global graph summary, enabling unsupervised learning of structurally aware node embeddings. Graph debiased contrastive learning \cite{zhao2021graph}  mitigates false negatives by leveraging clustering-derived pseudo-labels to sample negatives exclusively from distinct clusters, jointly refining representations and cluster assignments. Robust methods like RNCGLN \cite{zhu2024robust} further address dual noise (graph and label) via graph contrastive loss and self-attention, leveraging pseudo-graphs and pseudo-labels to correct supervision signals. GRLC \cite{peng2023grlc} extends this paradigm by maximizing mutual information between semantic and structural information while introducing constraints—negative embedding weighting, structural updating, and loss bounds—to bridge the gap between contrastive learning and downstream task performance. 

These unsupervised objectives maximize consistency of graph topology and attributes. In contrast, GRAVITY operates in a fully supervised regime: label information guides its latent gravitational alignment so that nodes of the same class are drawn together according to structural similarity. Graph Transformers also capture global topology: for example, Graphormer \cite{ying2021transformers} adds centrality and distance encodings to Transformer attention; GraphGPS \cite{rampavsek2022recipe} decouples local message-passing from global attention; SAT \cite{chen2022structure} augments node embeddings with subgraph context; and recent methods project nodes onto diverse geometric manifolds \cite{jyothish2025leveraging}. These strategies use fixed positional or geometric cues to encode structure, whereas GRAVITY achieves global awareness through learned forces. In this way, GRAVITY effectively acts as an implicit structural–semantic regularizer built into supervised learning \cite{borzone2025hybrid}, aligning labels with graph patterns without needing separate auxiliary tasks.

\section{Conclusion}

GRAVITY introduces a novel physics-inspired paradigm for supervised vertex representation learning, modeling graphs as dynamic systems where vertices self-organize under class-guided gravitational forces. By simulating attraction between same-class vertices and repulsion across different classes, the framework induces a latent potential field that sharpens semantic boundaries and promotes geometric coherence in embeddings. Crucially, it overcomes limitations of static message-passing schemes through its adaptive force-driven aggregation mechanism, which dynamically modulates each vertex’s receptive field based on structural proximity and attribute similarity. This enables context-aware integration of both local and global topological cues, allowing the model to uncover latent substructures while maintaining scalability.  

Empirical validation across diverse benchmarks demonstrates GRAVITY’s superiority in inductive vertex classification, outperforming state-of-the-art GNNs such as GraphSAGE and GAT. Its force-based organization not only yields competitive accuracy but also generates visually distinct, class-aligned clusters in the embedding space—even for unseen graphs—validating the framework’s generalization capacity. By intrinsically aligning label semantics with graph topology through gravitational equilibrium, GRAVITY eliminates the need for auxiliary clustering objectives or complex augmentation strategies. Future work could extend this physics-inspired approach to heterogeneous graphs or explore applications in dynamic network settings.

\bibliography{references}

\end{document}